\documentclass[11pt,final,journal,onecolumn]{IEEEtran}
\usepackage{amsmath,amssymb,color,amsmath, graphicx, float, caption, subcaption, mathrsfs,color} \usepackage[boxed]{algorithm2e} 
\usepackage{bm}
\usepackage{epstopdf }
\usepackage{type1cm} 
\usepackage{lettrine}
\usepackage{cite}
\usepackage[font={footnotesize}]{caption}

\def\H{\bm{H}}
\def\I{\bm{I}}

\def\N{\bm{N}}

\def\e{\bm{e}}

\def\x{\bm{x}}
\def\y{\bm{y}}
\def\z{\bm{z}}

\def\argmin#1{\underset{#1}{\textrm{argmin}}}

\def\minim#1{\underset{#1}{\textrm{min}}}

\def\CRLB_pe{\text{CRLB}}
\def\CRLB_pe{\text{CRLB}(\mathbf{p}_{e})}

\def\N02{\frac{N_0}{2}}

\def\0{\mathbf{0}}
\def\1{\mathbf{1}}

\SetKw{Kw}{KwEvaluate}

\begin{document}
\title{Super-Resolution via Image-Adapted Denoising CNNs: Incorporating External and Internal Learning}
\author{
    Tom~Tirer,
    and Raja~Giryes \\

    \thanks{This work was supported by the European research council (ERC StG 757497 PI Giryes). 
The authors are with the School of Electrical Engineering, Tel Aviv University, Tel Aviv 69978, Israel. (email: tirer.tom@gmail.com, raja@tauex.tau.ac.il). Code will be available at https://github.com/tomtirer/IDBP-CNN-IA.}
          }
\maketitle

\begin{abstract}
While deep neural networks exhibit state-of-the-art results in the task of image super-resolution (SR) with a fixed known acquisition process (e.g., a bicubic downscaling kernel), they experience a huge performance loss when the real observation model mismatches the one used in training.
Recently, two different techniques suggested to mitigate this deficiency, i.e., enjoy the advantages of deep learning without being restricted by the training phase. The first one follows the plug-and-play (P\&P) approach that solves general inverse problems (e.g., SR) by using Gaussian denoisers for handling the prior term in model-based optimization schemes. 
The second builds on internal recurrence of information inside a single image, and trains a super-resolver network at test time on examples synthesized from the low-resolution image.
Our work incorporates these two independent strategies, enjoying the impressive generalization capabilities of deep learning, captured by the first, and further improving it through internal learning at test time. First, we apply a recent P\&P strategy to SR. Then, we show how it may become image-adaptive in test time. This technique outperforms the above two strategies on popular datasets and gives better results than other state-of-the-art methods in practical cases where the observation model is inexact or unknown in advance. 
\end{abstract}



\section{Introduction}
\label{sec1}

The problem of image Super-Resolution (SR) has been the focus of many deep learning works in the recent years, and has experienced increasing improvement in performance along with the developments in deep learning \cite{dong2014learning, Bruna16Super,kim2016accurate, wang2015self, ledig2017photo, lim2017enhanced, Sajjadi17Enhancenet, zhang2018image, 
 Wang_2018_CVPR_Workshops, Yang2018SR_Review}.
In fact, when the acquisition process of the low-resolution (LR) image is known and fixed (e.g. a bicubic downscaling kernel), Convolutional Neural Network (CNN) methods trained using the exact observation model clearly outperform other SR techniques, e.g. model-based optimization methods \cite{zhang2017learning,  dong2013nonlocally, egiazarian2015single, huang2015single}.

However, when there is a mismatch in the observation model between the training and test data the CNN methods exhibit significant performance loss \cite{zhang2017learning, ZSSR}. This behavior is certainly undesirable, because in real life the acquisition process is often inexact or unknown in advance. Therefore, several recent approaches have been proposed with the goal of enjoying the advantages of deep learning without being restricted by the assumptions made in training \cite{zhang2017learning, ZSSR,  zhang2018learning, ulyanov2017deep}.

One line of works relies on the Plug-and-Play (P\&P) approach, introduced in \cite{venkatakrishnan2013plug}, which suggests leveraging excellent performance of denoising algorithms for solving other inverse imaging problems that can be formulated as a typical cost function, composed of fidelity and prior terms. The P\&P approach uses iterative optimization schemes, where the fidelity term is handled by a relatively simple optimization process and the prior term is handled by applying Gaussian denoisers. Most of the P\&P works \cite{venkatakrishnan2013plug, sreehari2016plug, chan2017plug, zhang2017learning, sun2019online, ono2017primal} directly apply existing optimization methods, such as ADMM \cite{boyd2011distributed} and FISTA \cite{beck2009fast}, that include steps in which the proximal mapping of the prior is used (equivalent to Gaussian denoising). A few others, e.g. RED \cite{romano2017little} and IDBP \cite{tirer2018image}, start with modifying the prior term \cite{romano2017little, reehorst2019regularization} or the fidelity term \cite{tirer2018image} of the typical cost function.

While the P\&P approach is not directly connected to deep learning, a P\&P method termed IRCNN \cite{zhang2017learning} has presented impressive SR results using a set of CNN Gaussian denoisers, providing a way to enjoy the generalization capabilities of deep learning as a natural image prior without any restrictions on the observation model.

Another recent strategy which is not restricted by the offline training phase relies on internal recurrence of information inside a single image \cite{glasner2009super, zontak2011internal, huang2015single}. 
In the spirit of this phenomenon, the SR method in \cite{ZSSR}, termed ZSSR, completely avoids offline training. Instead, it trains from scratch a super-resolver CNN at test time on examples synthesized from the LR image using an input blur kernel. 
This method relates to deep image prior \cite{ulyanov2017deep}, another deep learning solution for inverse imaging problems that optimizes the weights of a deep neural network only in the test phase.

{\bf Contribution.} In this paper we incorporate the two independent strategies mentioned above, enjoying the impressive generalization capabilities of deep learning, captured by the first, and further improving it by internal learning at test time. We start with the recently proposed IDBP framework \cite{tirer2018image}, which has been applied so far only to inpainting and deblurring. 
Here we apply it to SR using a set of CNN denoisers (same as those used by IRCNN \cite{zhang2017learning}) and obtain very good results. This IDBP-based SR method serves us as a strong starting point. 
We propose to further improve the performance by fine-tuning its CNN denoisers in test time using the LR input and synthetic additive Gaussian noise.

\begin{figure*}
\captionsetup[subfigure]{labelformat=empty}
  \centering

  \begin{subfigure}[b]{0.11\linewidth}
    \centering\includegraphics[width=35pt]{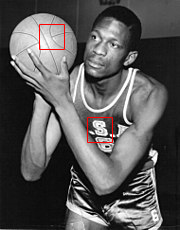}\\
    \caption{LR image}
\vspace{2mm}
  \end{subfigure}%
  \begin{subfigure}[b]{0.15\linewidth}
    \centering\includegraphics[width=70pt]{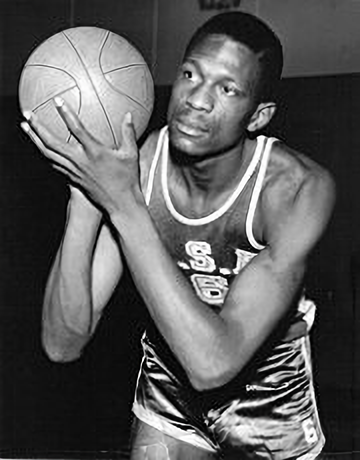}\\
\vspace{1mm}
    \centering\includegraphics[width=33pt]{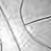}
    \centering\includegraphics[width=33pt]{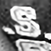}
    \caption{EDSR+ \cite{lim2017enhanced}}
\vspace{2mm}
  \end{subfigure}
  \begin{subfigure}[b]{0.15\linewidth}
    \centering\includegraphics[width=70pt]{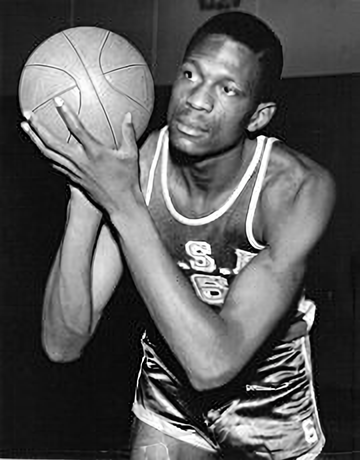}\\
\vspace{1mm}
    \centering\includegraphics[width=33pt]{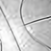}
    \centering\includegraphics[width=33pt]{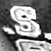}
    \caption{RCAN \cite{zhang2018image}}
\vspace{2mm}
  \end{subfigure}
  \begin{subfigure}[b]{0.15\linewidth}
    \centering\includegraphics[width=70pt]{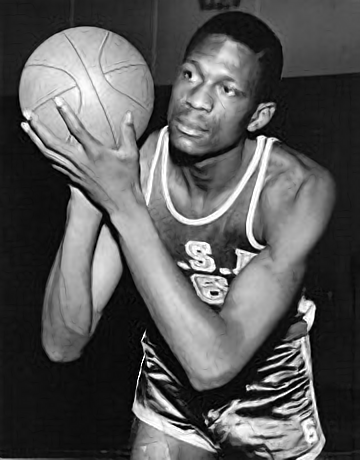}\\
\vspace{1mm}
    \centering\includegraphics[width=33pt]{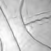}
    \centering\includegraphics[width=33pt]{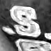}
    \caption{ZSSR \cite{ZSSR}}
\vspace{2mm}
  \end{subfigure}%
  \begin{subfigure}[b]{0.15\linewidth}
    \centering\includegraphics[width=70pt]{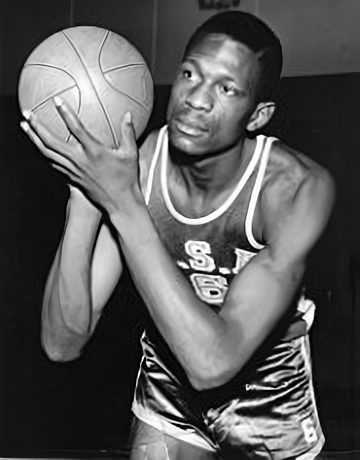}\\
\vspace{1mm}
    \centering\includegraphics[width=33pt]{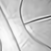}
    \centering\includegraphics[width=33pt]{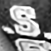}
    \caption{IDBP-CNN}
\vspace{2mm}
  \end{subfigure}
  \begin{subfigure}[b]{0.15\linewidth}
    \centering\includegraphics[width=70pt]{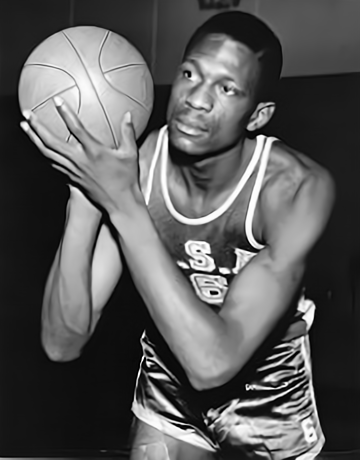}\\
\vspace{1mm}
    \centering\includegraphics[width=33pt]{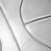}
    \centering\includegraphics[width=33pt]{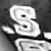}
    \caption{IDBP-CNN-IA}
\vspace{2mm}
  \end{subfigure}

  \begin{subfigure}[b]{0.11\linewidth}
    \centering\includegraphics[width=35pt]{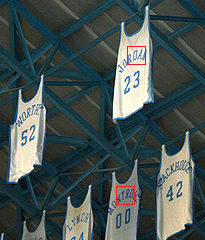}\\
    \caption{LR image}
\vspace{2mm}
  \end{subfigure}%
  \begin{subfigure}[b]{0.15\linewidth}
    \centering\includegraphics[width=70pt]{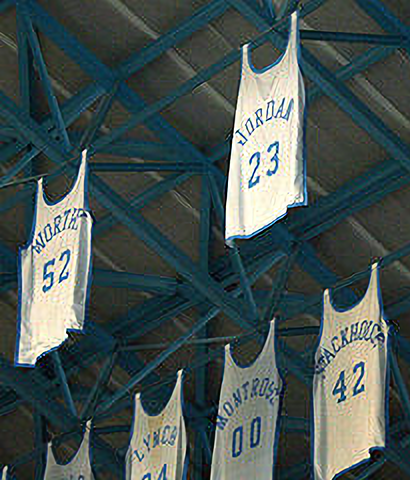}\\
\vspace{1mm}
    \centering\includegraphics[width=33pt]{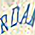}
    \centering\includegraphics[width=33pt]{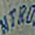}
    \caption{EDSR+ \cite{lim2017enhanced}}
\vspace{2mm}
  \end{subfigure}
  \begin{subfigure}[b]{0.15\linewidth}
    \centering\includegraphics[width=70pt]{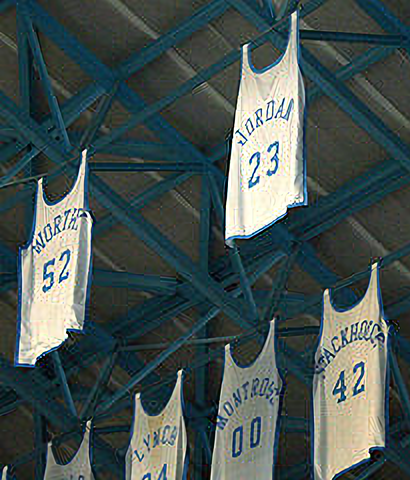}\\
\vspace{1mm}
    \centering\includegraphics[width=33pt]{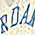}
    \centering\includegraphics[width=33pt]{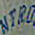}
    \caption{RCAN \cite{zhang2018image}}
\vspace{2mm}
  \end{subfigure}
  \begin{subfigure}[b]{0.15\linewidth}
    \centering\includegraphics[width=70pt]{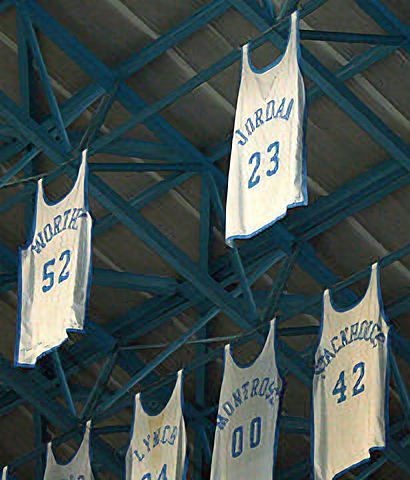}\\
\vspace{1mm}
    \centering\includegraphics[width=33pt]{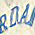}
    \centering\includegraphics[width=33pt]{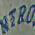}
    \caption{ZSSR \cite{ZSSR}}
\vspace{2mm}
  \end{subfigure}%
  \begin{subfigure}[b]{0.15\linewidth}
    \centering\includegraphics[width=70pt]{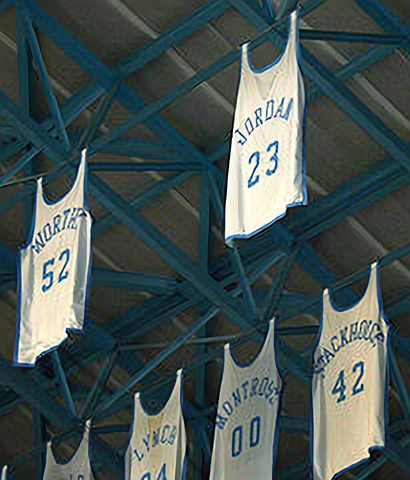}\\
\vspace{1mm}
    \centering\includegraphics[width=33pt]{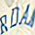}
    \centering\includegraphics[width=33pt]{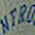}
    \caption{IDBP-CNN}
\vspace{2mm}
  \end{subfigure}
  \begin{subfigure}[b]{0.15\linewidth}
    \centering\includegraphics[width=70pt]{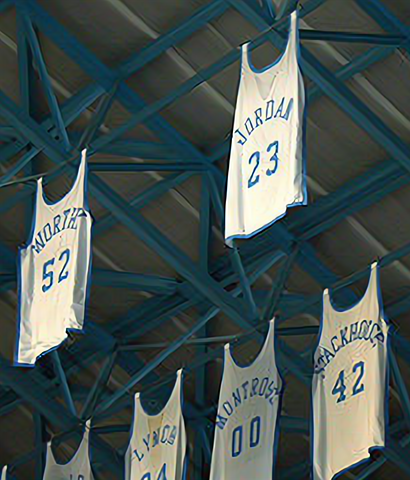}\\
\vspace{1mm}
    \centering\includegraphics[width=33pt]{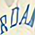}
    \centering\includegraphics[width=33pt]{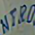}
    \caption{IDBP-CNN-IA}
\vspace{2mm}
  \end{subfigure}


  \begin{subfigure}[b]{0.11\linewidth}
    \centering\includegraphics[width=35pt]{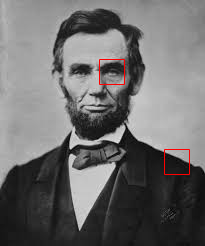}\\
    \caption{LR image}
\vspace{2mm}
  \end{subfigure}%
  \begin{subfigure}[b]{0.15\linewidth}
    \centering\includegraphics[width=70pt]{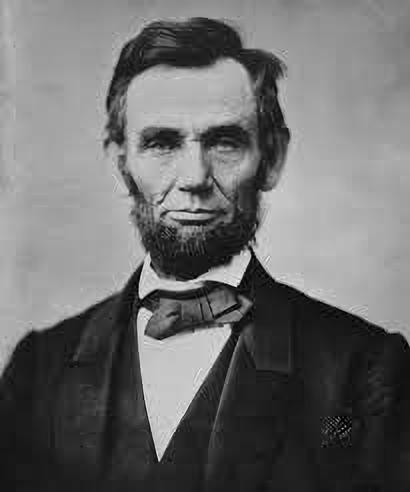}\\
\vspace{1mm}
    \centering\includegraphics[width=33pt]{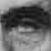}
    \centering\includegraphics[width=33pt]{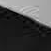}
    \caption{EDSR+ \cite{lim2017enhanced}}
\vspace{2mm}
  \end{subfigure}
  \begin{subfigure}[b]{0.15\linewidth}
    \centering\includegraphics[width=70pt]{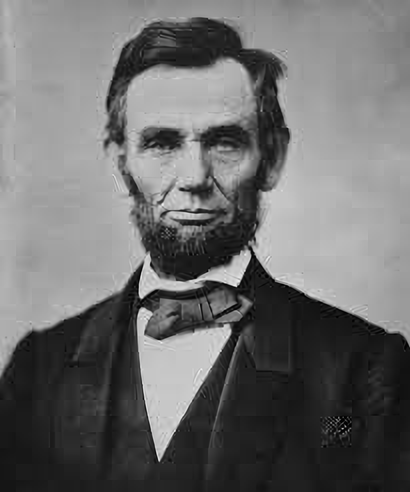}\\
\vspace{1mm}
    \centering\includegraphics[width=33pt]{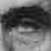}
    \centering\includegraphics[width=33pt]{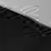}
    \caption{RCAN \cite{zhang2018image}}
\vspace{2mm}
  \end{subfigure}
  \begin{subfigure}[b]{0.15\linewidth}
    \centering\includegraphics[width=70pt]{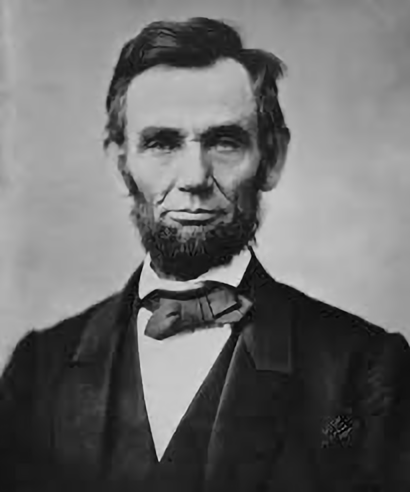}\\
\vspace{1mm}
    \centering\includegraphics[width=33pt]{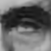}
    \centering\includegraphics[width=33pt]{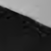}
    \caption{ZSSR \cite{ZSSR}}
\vspace{2mm}
  \end{subfigure}%
  \begin{subfigure}[b]{0.15\linewidth}
    \centering\includegraphics[width=70pt]{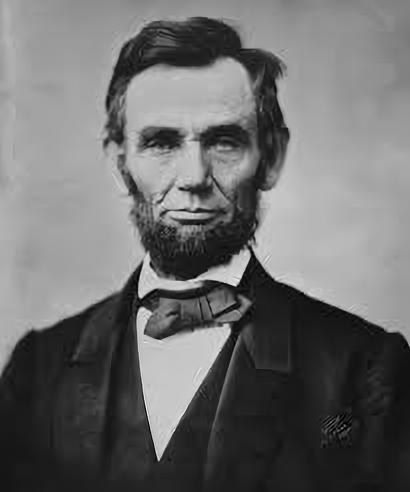}\\
\vspace{1mm}
    \centering\includegraphics[width=33pt]{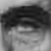}
    \centering\includegraphics[width=33pt]{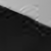}
    \caption{IDBP-CNN}
\vspace{2mm}
  \end{subfigure}
  \begin{subfigure}[b]{0.15\linewidth}
    \centering\includegraphics[width=70pt]{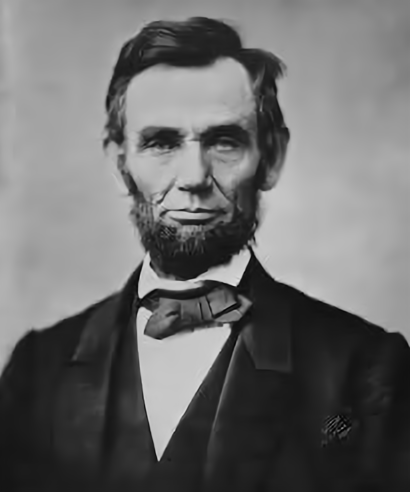}\\
\vspace{1mm}
    \centering\includegraphics[width=33pt]{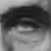}
    \centering\includegraphics[width=33pt]{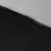}
    \caption{IDBP-CNN-IA}
\vspace{2mm}
  \end{subfigure}

  \caption{SR(x2) of old real images. We introduce a super-resolution version of the IDBP framework \cite{tirer2018image} that uses CNN denoisers and performs SR for any given down-sampling operator without retraining. We show that by making the CNN denoisers image-adaptive (IA), we get a more accurate image reconstruction with less artifacts. 
More examples (in larger size) are presented in Figs. \ref{real_example1}-\ref{real_example7}.
  }
\label{real_example}
\end{figure*}

Our image-adaptive approach improves over the plain IDBP method, which does not use any internal learning, as well as over ZSSR that uses only internal learning, on widely-used datasets and experiments. 
In practical cases, where the observation model is inexact or unknown in advance,  
it also gives better results than EDSR+ \cite{lim2017enhanced} and RCAN \cite{zhang2018image}, recent state-of-the-art SR methods.
Fig. \ref{real_example} presents the results obtained for several ''imperfect'' real images. 

\section{Problem formulation and IDBP-based SR}
\label{Sec:2}

Many image acquisition models can be formulated by 
\begin{align}
\label{Eq_general_model}
\y = \H\x + \e,
\end{align}
where $\x \in \Bbb R^n$ represents the unknown original image, $\y \in \Bbb R^m$ represents the observations, $\H$ is an $m \times n$ degradation matrix and  $\e \sim \mathcal{N}(\0,\sigma_e^2 \I_m)$. 
Using certain structures of $\H$, (\ref{Eq_general_model}) 
can be used for tasks like denoising, inpainting, deblurring, and more.
Specifically, here we are interested in image super-resolution, where $\H$ is a composite operator of blurring (e.g. anti-aliasing filtering) and down-sampling (hence $m < n$).

Most of the model-based approaches for recovering $\x$, try to solve a typical optimization problem composed of fidelity and prior terms
\begin{align}
\label{Eq_cost_func1}
\minim{\tilde{\x}} \,\,\, \frac{1}{2\sigma_e^2} \| \y-\H\tilde{\x} \|_2^2 + s(\tilde{\x}),
\end{align}
where $\tilde{\x}$ is the optimization variable, $\| \cdot \|_2$ stands for the Euclidean norm, and $s(\tilde{\x})$ is a prior model.
Recently, the work in \cite{tirer2018image} has suggested to solve a different optimization problem
\begin{align}
\label{Eq_cost_func_our}
\minim{\tilde{\x}, \tilde{\z}} \,\,\, \frac{1}{2(\sigma_e+\delta)^2} \| \tilde{\z}-\tilde{\x} \|_2^2 + s(\tilde{\x}) \,\,\,\, \textrm{s.t.} \,\,\,\, \H\tilde{\z}= \y,
\end{align}
where $\delta$ is a design parameter that should be set according to a certain condition that keeps \eqref{Eq_cost_func_our} as an approximation of \eqref{Eq_cost_func1} (see Section III in \cite{tirer2018image} for more details).
The major advantage of \eqref{Eq_cost_func_our} over \eqref{Eq_cost_func1} is the possibility to solve it using a simple alternating minimization scheme that possesses the plug-and-play property: the prior term $s(\tilde{\x})$ is handled solely by a Gaussian denoising operation $\mathcal{D}(\cdot;\sigma)$ with noise level $\sigma=\sigma_e+\delta$. Iteratively, $\tilde{\x}_k$ is obtained by 
\begin{align}
\label{Eq_cost_func_our_x}
\tilde{\x}_k &= \argmin{\tilde{\x}} \,\, \frac{1}{2(\sigma_e+\delta)^2} \| \tilde{\z}_{k-1}-\tilde{\x} \|_2^2 + s(\tilde{\x}) \nonumber \\
&=\mathcal{D}(\tilde{\z}_{k-1};\sigma_e+\delta),
\end{align}
and $\tilde{\z}_k$ is obtained by projecting $\tilde{\x}_k$ onto $\{ \H \Bbb R^n = \y \}$
\begin{align}
\label{Eq_cost_func_our_y}
\tilde{\z}_k &= \argmin{\tilde{\z}} \,\, \| \tilde{\z}-\tilde{\x}_k \|_2^2  \,\,\,\, \textrm{s.t.} \,\,\,\, \H\tilde{\z}= \y \nonumber  \\
&= \H^\dagger\y + (\I_n - \H^\dagger\H)\tilde{\x}_k \nonumber  \\
&= \H^\dagger (\y - \H \tilde{\x}_k) + \tilde{\x}_k,
\end{align}
where $\H^\dagger \triangleq \H^T(\H\H^T)^{-1}$ is the pseudoinverse of $\H$ (recall $m < n$). The two repeating operations lends the method its name: Iterative Denoising and Backward Projections (IDBP). After a stopping criterion is met, the last $\tilde{\x}_k$ is taken as the estimate of the latent image $\x$.

The IDBP method can be applied to SR in an efficient manner: the composite operators $\H$ and $\H^T$ are easy to perform, and matrix inversion can be avoided using the conjugate gradient method. We note that until now IDBP performance has been demonstrated only for inpainting and deblurring \cite{tirer2018image, tirer2018iterative}.
Moreover, contrary to prior work, here we adopt the strategy in \cite{zhang2017learning} of changing CNN denoisers during the iterations. 
We denote the resulting method by IDBP-CNN. 


\begin{figure}
  \centering
  \begin{subfigure}[b]{0.5\linewidth}
    \centering\includegraphics[width=0.8\linewidth]{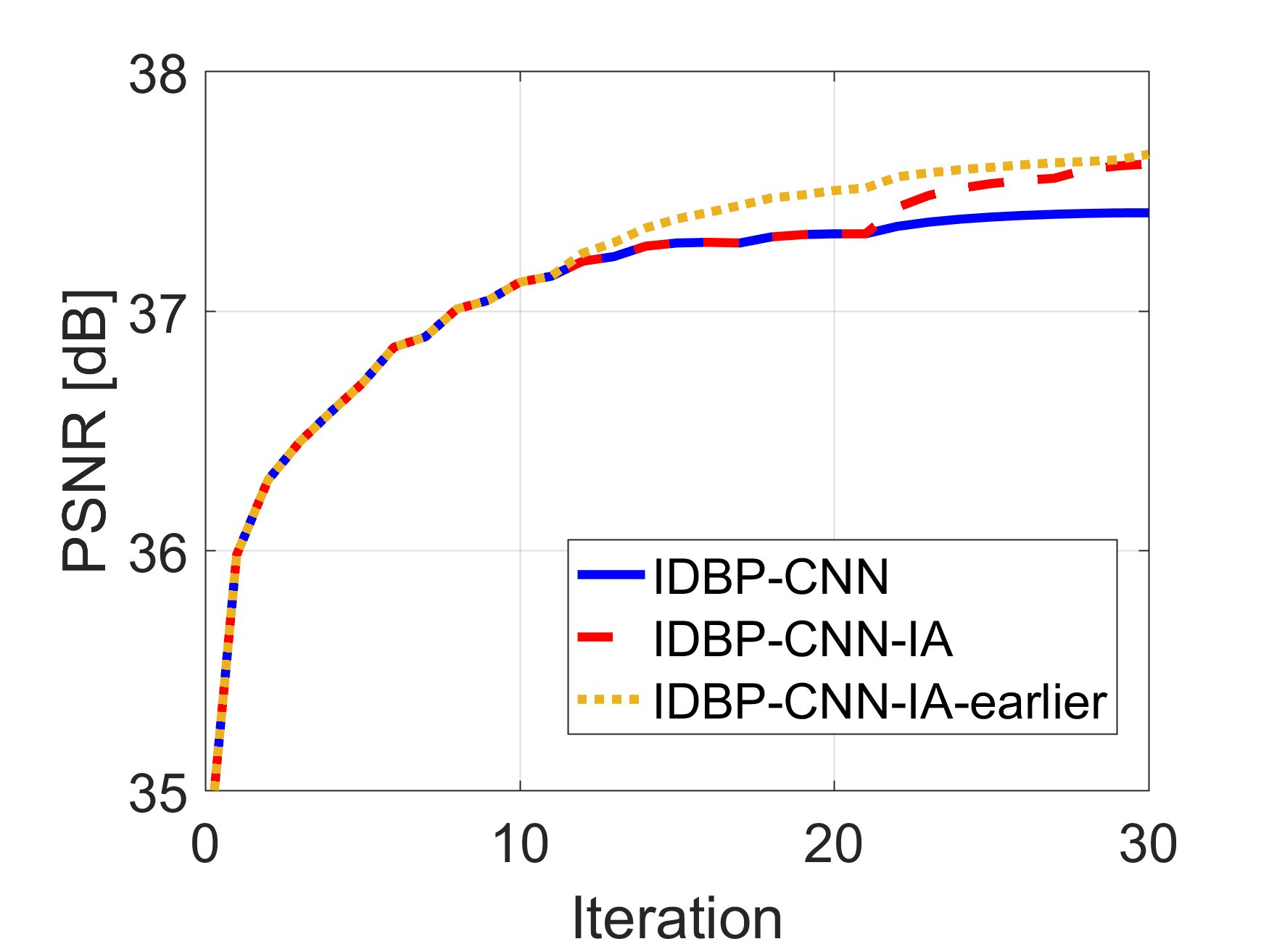}
    \caption{\label{fig:psnr_vs_iter_bicub}}
  \end{subfigure}%
  \begin{subfigure}[b]{0.5\linewidth}
    \centering\includegraphics[width=0.8\linewidth]{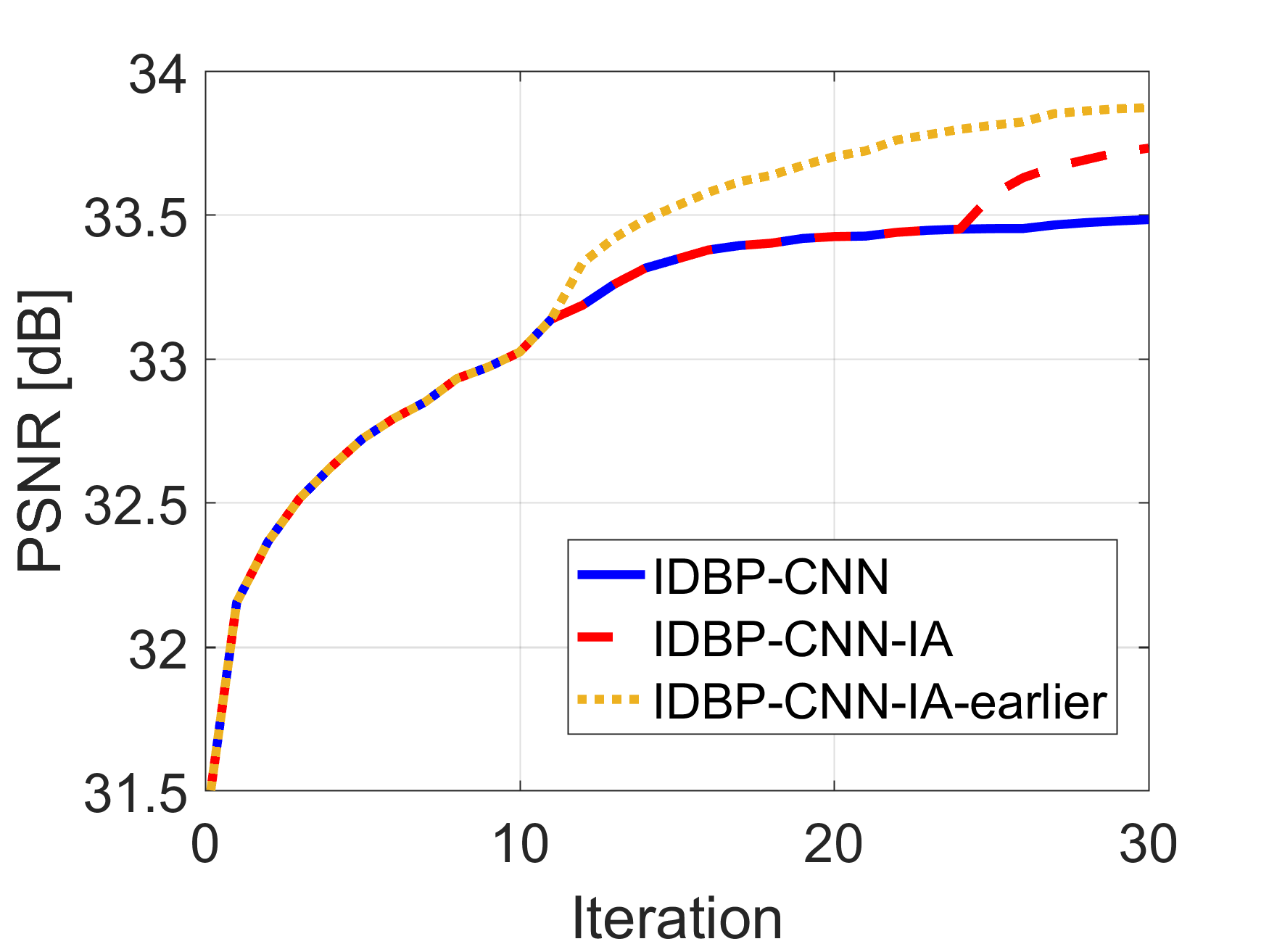}
    \caption{\label{fig:psnr_vs_iter_gauss}}
  \end{subfigure}
    \caption{Super-resolution results (PSNR averaged on Set5 vs. iteration number) for IDBP-CNN with and without our image-adapted CNN approach: (\subref{fig:psnr_vs_iter_bicub}) SR x2 with bicubic kernel; (\subref{fig:psnr_vs_iter_gauss}) SR x3 with Gaussian kernel. A boost in performance is observed once the IDBP scheme starts using image-adapted CNN denoisers.} 
\label{fig:PSNR_vs_iter}
\end{figure}

\subsection{IDBP-based SR with Image-adapted CNNs}
\label{Sec:Image-adapted}

As described above, the P\&P approach allows to use CNN denoisers that handle only the prior term in the iterative scheme. Therefore, no assumptions on the observation model are done in the offline training phase.
Yet, we suggest to have an {\em additional ingredient} \textemdash an internal learning step, where the CNN denoisers are fine-tuned in test-time using the LR input. 
This way we obtain {\em image-adapted} CNN denoisers that can perform better on patterns that are specific to this image and even better remove artifacts that do not recur in fixed patterns. 

When the observed LR image does not exhibit any degradation (e.g. additive noise, compression artifacts, blur), 
the phenomenon of recurrence of patterns within and across scales \cite{glasner2009super} implies that information inside the LR can improve the CNN denoiser compared to using only prior training on external data. 
When the quality of the LR image reduces, 
the achievable improvement is expected to decrease. However, as we show below, it is still possible to capture useful information from the LR image even when it suffers from a moderate degradation (e.g. in old images) or blurred with an unknown blur. 
The latter can be handled 
by using a common kernel estimation method that supplies as a by-product an intermediate estimation of the image with enhanced edges and textures (e.g. see the introduction section in \cite{tirer2018iterative}).

We note that several recent works demonstrate performance improvement of denoisers if they are learned or fine-tuned in the training phase using a set of images from the same class as the desired image \cite{teodoro2016image,remez2018class}. In contrast, here we fine-tune CNN denoisers in test-time using a single LR observation.

\subsection{Implementation}
\label{sec:Implementation}

We use for IDBP-CNN the same set of grayscale\footnote{We apply our method on the luminance channel, and use simple bicubic upsampling to obtain the color channels. However, our method can be extended in a straightforward way by using color denoisers.} CNN denoisers that are proposed and trained in \cite{zhang2017learning}. This set is composed of 25 CNNs, each of them is trained for a different noise level, and together they span the noise level range of $[0, 50]$. The CNNs architecture is presented in \cite{zhang2017learning}.


The IDBP-CNN uses a fixed number of 30 iterations, in which it alternates between \eqref{Eq_cost_func_our_y} and \eqref{Eq_cost_func_our_x}, where $\tilde{\x}_0$ is initialized using bicubic upsampling. The value of $\delta$ in \eqref{Eq_cost_func_our_x} is reduced exponentially from $12s$ to $s$, where $s$ denotes the desired SR scale factor. 
In most of the experiments in this paper $\sigma_e=0$. In this case, as discussed in \cite{tirer2018image} for noiseless inpainting, IDBP theory allows to decrease $\delta$ to any small positive value as the iterations increase. However, in experiments with $\sigma_e>0$ (LR with additive degradation) we set a fixed lower bound on the value of $\delta$ 
to ensure good performance. 
Let us denote the monotonically decreasing sequence of $\delta$ by $\{\delta_k\}$. In each iteration, a suitable CNN denoiser (i.e. associated with $\sigma_e+\delta_k$) is used. After 30 iterations, an estimator of the high-resolution image $\x$ is obtained by the last $\tilde{\x}_k$ 
(or $\tilde{\z}_k$ in the case of $\sigma_e=0$, as done for noiseless inpainting in \cite{tirer2018image}).

We now turn to discuss the implementation of our image-adaptive method, which we denote by IDBP-CNN-IA. In order to examine the effect of this idea, we use the same IDBP-CNN algorithm with a single change: the CNN denoisers are obtained by fine-tuning the pre-trained denoisers using the LR image. 
The fine-tuning of a CNN denoiser is done as follows. We extract patches of size uniformly chosen from $\{34, 40, 50\}$ from the LR image $\y$, which serve as the ''ground truth''. Their noisy version are obtained by adding random Gaussian noise of the level that is used in the offline training.
To enrich this ''training set'', data augmentation is done by downscaling $\y$ to 0.9 of its size with probability 0.5, using mirror reflections in the vertical and horizontal directions with uniform probability, and using 4 rotations $\{0^{\circ}, 90^{\circ}, 180^{\circ}, 270^{\circ}\}$, again, with uniform probability.
The optimization process, which is done in test-time, is kept fast and simple. We use L1 loss (we use residual learning as done in the training phase \cite{zhang2017learning}), minibatch size of 32, and 320 iterations of ADAM optimizer \cite{kingma2014adam} with its default parameters and learning rate of 3e-4. 

The fine-tuning time for a single denoiser is small and independent of the image size and the desired SR scale-factor. 
However, if the fine-tuning is done for every denoiser, the inference run-time becomes very large. For this reason, in our reported results we fine-tune only the last two CNN denoisers, and thus only moderately increase the inference run-time compared to the baseline IDBP-CNN. Due to the exponential decrease of $\{\delta_k\}$, these two denoisers are used in many last iterations (5-8 iterations out of 30). 

In Fig. \ref{fig:PSNR_vs_iter} we present the PSNR, averaged on Set5 dataset, as a function of the iteration number for IDBP-CNN with and without our IA approach, for two observation models. 
In both scenarios, a boost in performance is observed once the IDBP scheme starts using the fine-tuned CNN denoisers. We also present the results obtained by start using IA at an earlier iteration (i.e. fine-tuning also denoisers of higher noise levels). Such configuration, which significantly increases run-time, improves the results (as expected), 
but not always significantly. Presumably, because the high-level denoisers in early iterations improve mainly coarse details.


\begin{table*}
\scriptsize 
\renewcommand{\arraystretch}{1.3}
\caption{Super-resolution results (average PSNR in dB) for ideal (noiseless) observation model with bicubic and Gaussian downscaling kernels. Bold black indicates the leading method, and bold blue indicates the leading model-flexible method.} 
\label{table:results1}
\centering
    \begin{tabular}{ | c | c | c || c | c | c | c | c | c | c | c |}
    \hline
 Dataset  & Scale & Kernel & SRCNN \cite{dong2014learning} & VDSR \cite{kim2016accurate} & EDSR+ \cite{lim2017enhanced} & RCAN \cite{zhang2018image} & IRCNN \cite{zhang2017learning} & ZSSR \cite{ZSSR} & IDBP-CNN & IDBP-CNN-IA \\ \hline
    Set5 & \begin{tabular}{c}2 \\ 3 \\ 3 \end{tabular} & \begin{tabular}{c}Bicubic \\ Bicubic \\ Gaussian \end{tabular} & \begin{tabular}{c}36.66 \\ 32.75 \\ 30.42 \end{tabular} & \begin{tabular}{c}37.53 \\ 33.66 \\ 30.54 \end{tabular} & \begin{tabular}{c} 38.20 \\ {\bf 34.76} \\ 30.65 \end{tabular} & \begin{tabular}{c}{\bf 38.27} \\ 34.74 \\ 30.74 \end{tabular} & \begin{tabular}{c}37.43 \\ 33.39 \\ 33.38 \end{tabular} & \begin{tabular}{c}37.37 \\ 33.42 \\ 31.31 \end{tabular}  & \begin{tabular}{c}37.41 \\ 33.44 \\ 33.48 \end{tabular} & \begin{tabular}{c} \textcolor{blue}{{\bf 37.62}} \\ \textcolor{blue}{{\bf 33.60}} \\ {\bf 33.73} \end{tabular} \\ \hline
    Set14 & \begin{tabular}{c}2 \\ 3 \\ 3 \end{tabular} & \begin{tabular}{c}Bicubic \\ Bicubic \\ Gaussian \end{tabular} & \begin{tabular}{c}32.42 \\ 29.28 \\ 27.71 \end{tabular} & \begin{tabular}{c}33.03 \\ 29.77 \\ 27.80 \end{tabular}	& \begin{tabular}{c} 34.02 \\ {\bf 30.66} \\ 27.54 \end{tabular} & \begin{tabular}{c}{\bf 34.12} \\ 30.65 \\ 27.80 \end{tabular} & \begin{tabular}{c}32.88 \\ 29.61 \\ 29.63 \end{tabular} & \begin{tabular}{c}33.00 \\ \textcolor{blue}{{\bf  29.80}} \\ 28.33 \end{tabular} & \begin{tabular}{c}32.95 \\ 29.65 \\ 29.68 \end{tabular} & \begin{tabular}{c}\textcolor{blue}{{\bf  33.09}} \\ 29.72 \\ {\bf 29.79} \end{tabular} \\ \hline
    BSD100 & \begin{tabular}{c}2 \\ 3 \\ 3 \end{tabular} & \begin{tabular}{c}Bicubic \\ Bicubic \\ Gaussian \end{tabular}  & \begin{tabular}{c}31.36 \\ 28.41 \\ 27.32 \end{tabular}  & \begin{tabular}{c}31.90 \\ 28.82 \\ 27.43 \end{tabular} & \begin{tabular}{c} 32.37 \\ {\bf 29.32} \\ 27.46 \end{tabular} & \begin{tabular}{c}{\bf 32.41} \\ {\bf 29.32} \\ 27.52 \end{tabular} & \begin{tabular}{c}31.68 \\ 28.62 \\ 28.64 \end{tabular} & \begin{tabular}{c}31.65 \\ 28.67 \\ 27.76 \end{tabular} & \begin{tabular}{c}31.71 \\ 28.63 \\ 28.67 \end{tabular} & \begin{tabular}{c}\textcolor{blue}{{\bf 31.81}} \\ \textcolor{blue}{{\bf 28.68}} \\ {\bf 28.74} \end{tabular} \\ \hline
    \end{tabular}
\end{table*}

\section{Experiments}
\label{Sec:Experiments}

\subsection{''Ideal'' observation model}

In this section we assume known anti-aliasing kernels and no noise. 
We examine three cases: bicubic kernel with down-scaling factors of 2 and 3, and Gaussian kernel of size $7 \times 7$ with standard deviation 1.6 with down-scaling factor of 3. The latter scenario is used in many works \cite{dong2013nonlocally, romano2017little, zhang2017learning}.

We compare the IDBP-CNN with and without our image-adapted CNN approach to SRCNN \cite{dong2014learning}, VDSR \cite{kim2016accurate} and recent state-of-the-art methods EDSR+ \cite{lim2017enhanced} and RCAN \cite{zhang2018image}. All these four methods require extensive offline training to handle any different model \eqref{Eq_general_model}, and their benchmarked versions are available for the bicubic kernel cases. The goal of examining them for the Gaussian kernel is to show their huge performance loss whenever their training phase does not use the right observation model. We also compare our results to IRCNN \cite{zhang2017learning} and ZSSR \cite{ZSSR}, which are flexible to changes in the observation model like our approach. 
The results are given in Table \ref{table:results1}. 
The PSNR is computed on Y channel, as done in all the previous benchmarks.

It can be seen that IDBP-CNN-IA outperforms all other model-flexible methods. 
It also obtains the overall best results in the Gaussian kernel case.
Regarding the inference run-time, our experiments are performed on Intel i7-7500U CPU and Nvidia GeForce GTX 950M GPU. The IDBP-CNN requires $\sim$20s per image. Its image-adapted version requires $\sim$100s, which is only a moderate increase and is significantly faster than ZSSR that requires $\sim$150s in its fastest version (which achieves lower PSNR than reported in the table). 

\begin{table}
\footnotesize 
\renewcommand{\arraystretch}{1.3}
\caption{Super-resolution results (average PSNR in dB) for 8 estimated (inexact) non-ideal downscaling kernels and scale factor of 2.} \label{table:results2}
\centering
    \begin{tabular}{ | c || c | c | c | c | c | c |}
    \hline
 Dataset  & EDSR+ & RCAN &  IRCNN & ZSSR & IDBP-CNN & IDBP-CNN-IA \\ \hline
    Set5 (5$\times$8) & 29.99 & 30.01 & 32.30 & 33.35 & 33.33 & {\bf 33.46}  \\ \hline
    Set14 (14$\times$8) &  27.45 & 27.48 & 29.24 & 29.30 & 29.89 & {\bf 29.96}  \\ \hline
    \end{tabular}
\end{table}

\subsection{Unknown downscaling kernels}
\label{Sec:poorLR}

\begin{figure}
  \centering
  \begin{subfigure}[b]{0.2\linewidth}
    \centering\includegraphics[width=0.3\linewidth]{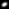} 
    \centering\includegraphics[width=0.3\linewidth]{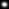}\\ \vspace{1mm}
    \centering\includegraphics[width=0.3\linewidth]{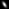}
    \centering\includegraphics[width=0.3\linewidth]{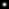}\\ \vspace{1mm}
    \centering\includegraphics[width=0.3\linewidth]{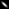}
    \centering\includegraphics[width=0.3\linewidth]{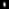}\\ \vspace{1mm}
    \centering\includegraphics[width=0.3\linewidth]{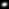}
    \centering\includegraphics[width=0.3\linewidth]{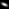}
    \caption{\label{fig:kernels}}
  \end{subfigure}%
  \begin{subfigure}[b]{0.9\linewidth}
    \centering\includegraphics[width=0.15\linewidth]{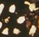} 
    \centering\includegraphics[width=0.15\linewidth]{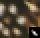}
    \centering\includegraphics[width=0.15\linewidth]{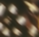}
    \centering\includegraphics[width=0.15\linewidth]{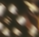}\\
\vspace{1mm}
    \centering\includegraphics[width=0.15\linewidth]{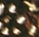}
    \centering\includegraphics[width=0.15\linewidth]{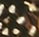}
    \centering\includegraphics[width=0.15\linewidth]{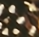}
    \caption{\label{fig:est_h_example}}
  \end{subfigure}
    \caption{(\subref{fig:psnr_vs_iter_bicub}) Non-ideal downscaling kernels; (\subref{fig:psnr_vs_iter_gauss}) SR x2 of {\em monarch}, Set5. From left to right and top to bottom, fragments of: original image, LR image with the estimated kernel, EDSR, RCAN, ZSSR, IDBP-CNN and IDBP-CNN-IA.} 
\label{fig:estH}
\end{figure}

In real life the downscaling kernel is often unknown and non-ideal (i.e. the LR may be blurry). To examine this situation, we use 8 different non-ideal kernels, as those used in \cite{ZSSR}, which are presented in Fig. \ref{fig:kernels}. Using these kernels we create 8 LR images for each ground truth image. Moreover, the {\em true kernels are assumed to be unknown}. Therefore, for the model-flexible methods (IDBP, IRCNN, ZSSR) we use the kernel estimation method proposed in \cite{sroubek2007unified} as an initial step. Similarly to previous section, EDSR+ and RCAN are restricted to the bicubic assumption made in their offline training phase. In fact, in this practical case, perhaps it is impossible to have a pure deep learning solution. Regarding the image-adaptive approach, instead of using the blurry LR for fine-tuning, we use its enhanced version (with strong edges and textures) which is a by-product of the kernel estimation algorithm. As mentioned above, it is very common that kernel estimation methods produce such images \cite{tirer2018iterative}. 

The results are given in Table \ref{table:results2}, and a visual example is shown in Fig. \ref{fig:est_h_example}. It can be seen that our IDBP-CNN performs well, and its IA version further improves it.

\subsection{Low-quality real LR images}
\label{Sec:oldLR}

In this section we visually compare the performance of the IDBP-based methods with ZSSR, EDSR+ and RCAN on real images (i.e. we do not have ground truth for their high-resolution versions). Specifically, we consider old images, which are somewhat degraded and whose acquisition model is unknown. Again, EDSR+ and RCAN cannot handle such images differently because they are restricted by the assumptions made in their training phase. For ZSSR we run the official code with its predefined configuration for handling real images. For IDBP-CNN we use a lower bound of $10$ on the values of $\{\delta_k\}$, i.e. when $\delta_k<10$ is reached, the IDBP scheme stops switching denoisers and keeps using the same denoiser for the rest of the 30 iterations. Again, for IDBP-CNN-IA we use the exact IDBP scheme except that we fine-tune this last CNN denoiser (and use it in the remaining iterations). 
We also note that all the examined methods assume bicubic kernel (while the true kernel of each image is unknown). 

\begin{figure*}
\captionsetup[subfigure]{labelformat=empty}
  \centering
  \begin{subfigure}[b]{0.48\linewidth}
    \centering\includegraphics[width=75pt]{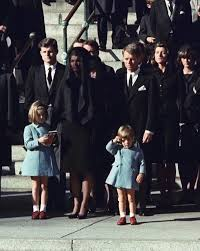}\\
    \caption{LR image}
\vspace{2mm}
  \end{subfigure}%
\\
  \begin{subfigure}[b]{0.3\linewidth}
    \centering\includegraphics[width=150pt]{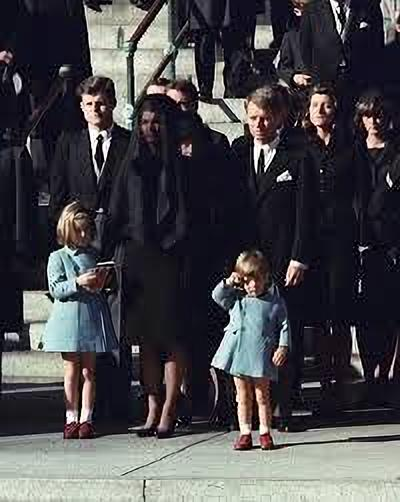}\\
    \caption{EDSR+}%
\vspace{2mm}
  \end{subfigure}%
  \begin{subfigure}[b]{0.3\linewidth}
    \centering\includegraphics[width=150pt]{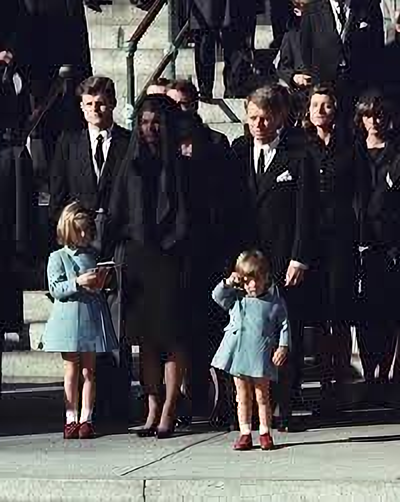}\\
    \caption{RCAN}%
\vspace{2mm}
  \end{subfigure}%
  \begin{subfigure}[b]{0.3\linewidth}
    \centering\includegraphics[width=150pt]{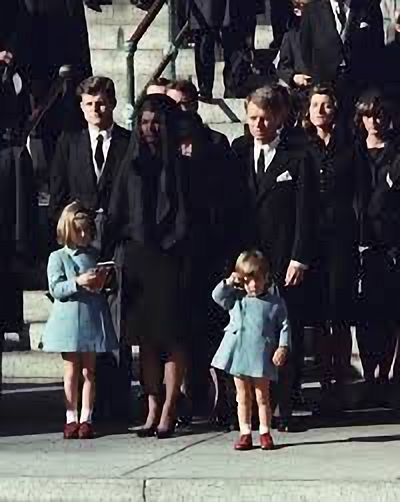}\\
    \caption{ZSSR}%
\vspace{2mm}
  \end{subfigure}%
\\
  \begin{subfigure}[b]{0.3\linewidth}
    \centering\includegraphics[width=150pt]{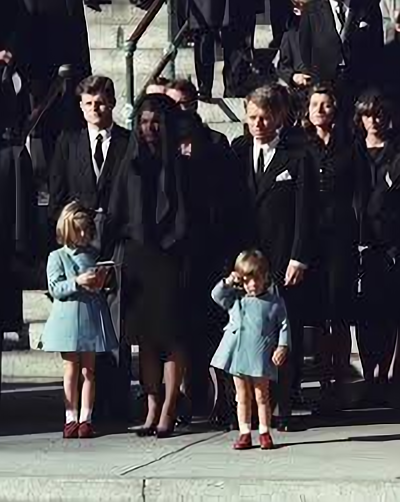}\\
    \caption{IDBP-CNN}%
\vspace{2mm}
  \end{subfigure}
  \begin{subfigure}[b]{0.3\linewidth}
    \centering\includegraphics[width=150pt]{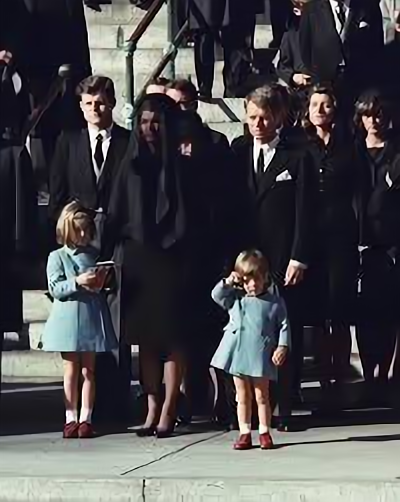}\\
    \caption{IDBP-CNN-IA}%
\vspace{2mm}
  \end{subfigure}
  \caption{SR(x2) of a real image. It is recommended to zoom at the images.}
\label{real_example1}
\end{figure*}

\begin{figure*}
\captionsetup[subfigure]{labelformat=empty}
  \centering
  \begin{subfigure}[b]{0.48\linewidth}
    \centering\includegraphics[width=75pt]{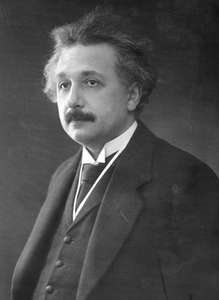}\\
    \caption{LR image}
\vspace{2mm}
  \end{subfigure}%
\\
  \begin{subfigure}[b]{0.3\linewidth}
    \centering\includegraphics[width=150pt]{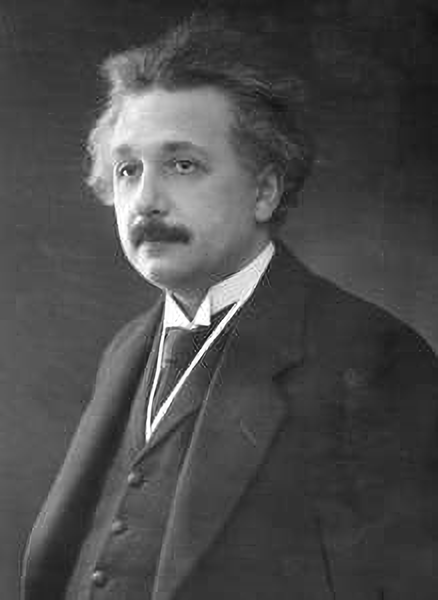}\\
    \caption{EDSR+}%
\vspace{2mm}
  \end{subfigure}%
  \begin{subfigure}[b]{0.3\linewidth}
    \centering\includegraphics[width=150pt]{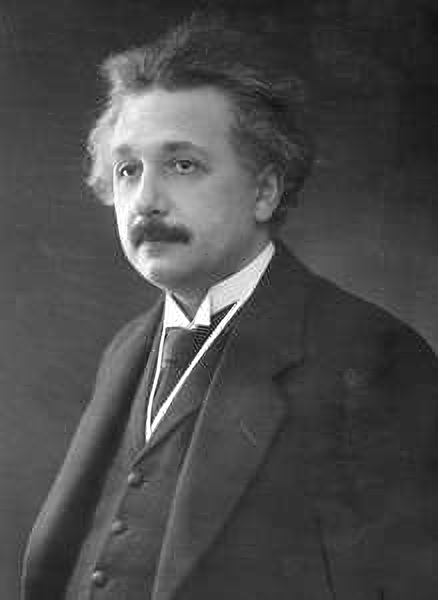}\\
    \caption{RCAN}%
\vspace{2mm}
  \end{subfigure}%
  \begin{subfigure}[b]{0.3\linewidth}
    \centering\includegraphics[width=150pt]{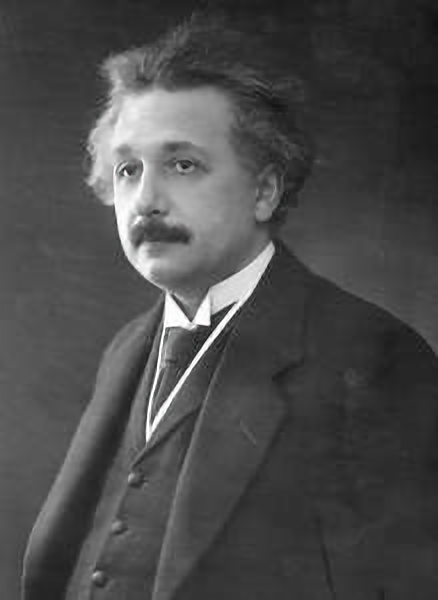}\\
    \caption{ZSSR}%
\vspace{2mm}
  \end{subfigure}%
\\
  \begin{subfigure}[b]{0.3\linewidth}
    \centering\includegraphics[width=150pt]{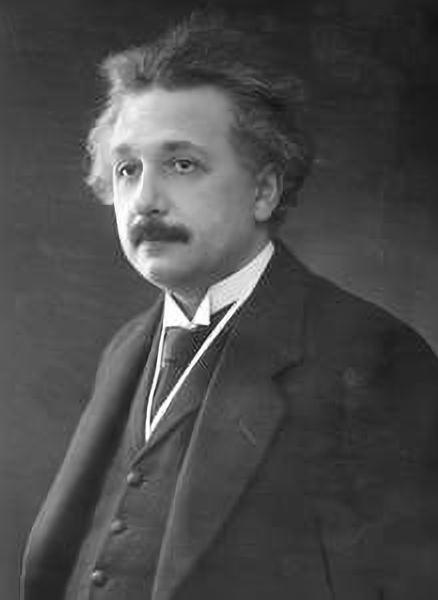}\\
    \caption{IDBP-CNN}%
\vspace{2mm}
  \end{subfigure}
  \begin{subfigure}[b]{0.3\linewidth}
    \centering\includegraphics[width=150pt]{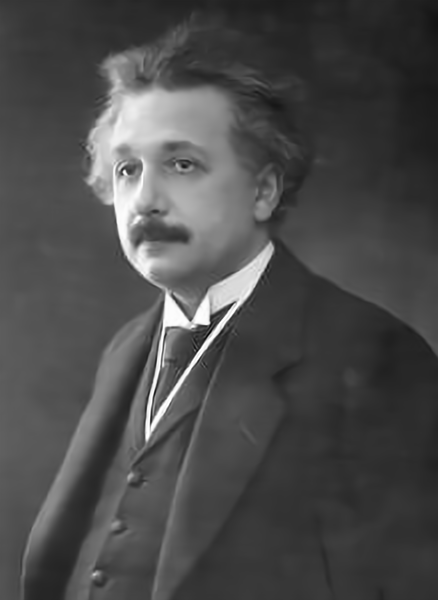}\\
    \caption{IDBP-CNN-IA}%
\vspace{2mm}
  \end{subfigure}
  \caption{SR(x2) of a real image. It is recommended to zoom at the images.}
\label{real_example2}
\end{figure*}

\begin{figure*}
\captionsetup[subfigure]{labelformat=empty}
  \centering
  \begin{subfigure}[b]{0.48\linewidth}
    \centering\includegraphics[width=75pt]{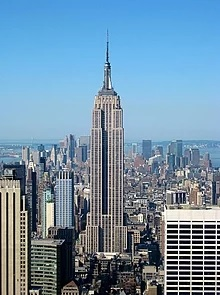}\\
    \caption{LR image}
\vspace{2mm}
  \end{subfigure}%
\\
  \begin{subfigure}[b]{0.3\linewidth}
    \centering\includegraphics[width=150pt]{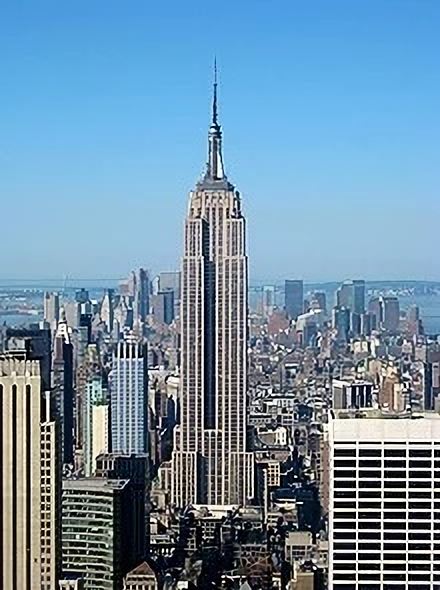}\\
    \caption{EDSR+}%
\vspace{2mm}
  \end{subfigure}%
  \begin{subfigure}[b]{0.3\linewidth}
    \centering\includegraphics[width=150pt]{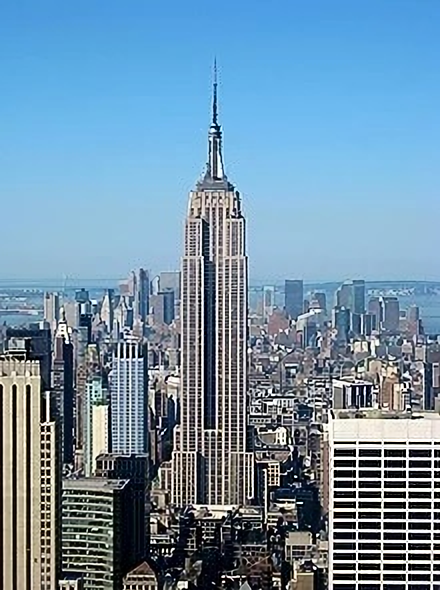}\\
    \caption{RCAN}%
\vspace{2mm}
  \end{subfigure}%
  \begin{subfigure}[b]{0.3\linewidth}
    \centering\includegraphics[width=150pt]{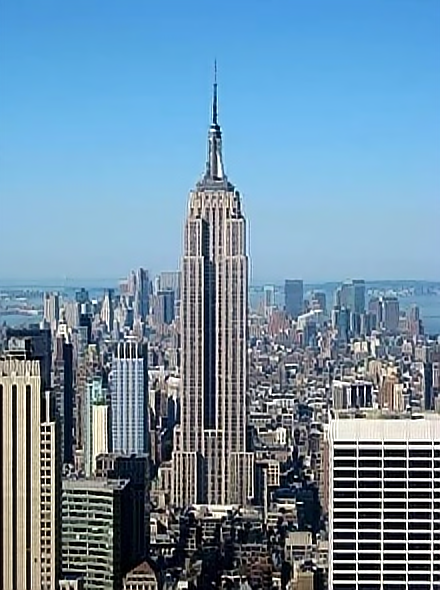}\\
    \caption{ZSSR}%
\vspace{2mm}
  \end{subfigure}%
\\
  \begin{subfigure}[b]{0.3\linewidth}
    \centering\includegraphics[width=150pt]{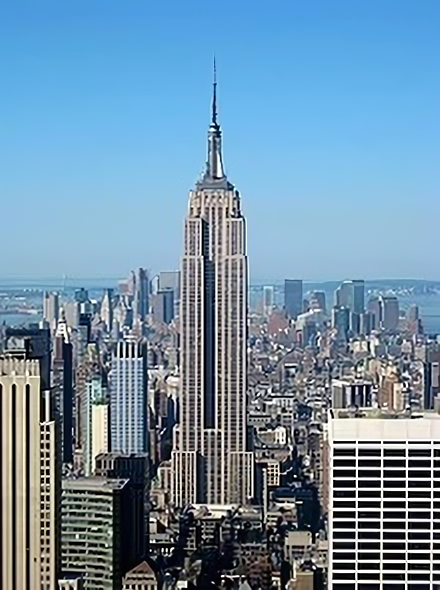}\\
    \caption{IDBP-CNN}%
\vspace{2mm}
  \end{subfigure}
  \begin{subfigure}[b]{0.3\linewidth}
    \centering\includegraphics[width=150pt]{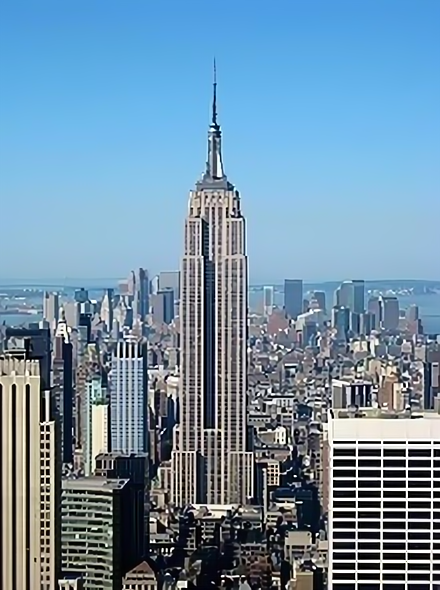}\\
    \caption{IDBP-CNN-IA}%
\vspace{2mm}
  \end{subfigure}
  \caption{SR(x2) of a real image. It is recommended to zoom at the images.}
\label{real_example3}
\end{figure*}

\begin{figure*}
\captionsetup[subfigure]{labelformat=empty}
  \centering
  \begin{subfigure}[b]{0.48\linewidth}
    \centering\includegraphics[width=75pt]{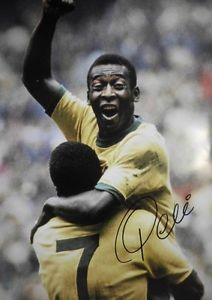}\\
    \caption{LR image}
\vspace{2mm}
  \end{subfigure}%
\\
  \begin{subfigure}[b]{0.3\linewidth}
    \centering\includegraphics[width=150pt]{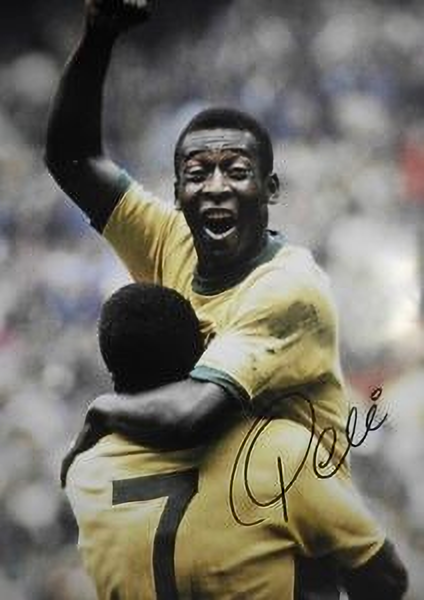}\\
    \caption{EDSR+}%
\vspace{2mm}
  \end{subfigure}%
  \begin{subfigure}[b]{0.3\linewidth}
    \centering\includegraphics[width=150pt]{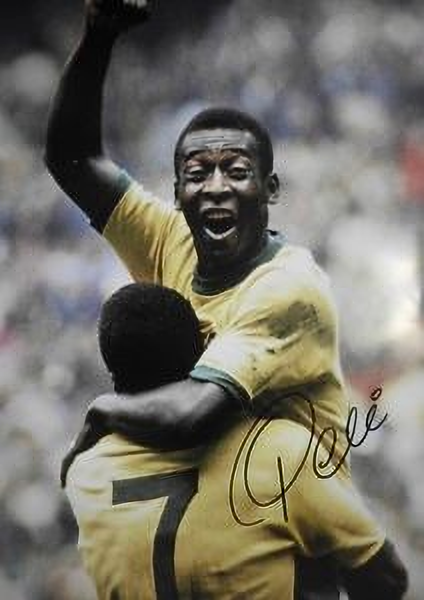}\\
    \caption{RCAN}%
\vspace{2mm}
  \end{subfigure}%
  \begin{subfigure}[b]{0.3\linewidth}
    \centering\includegraphics[width=150pt]{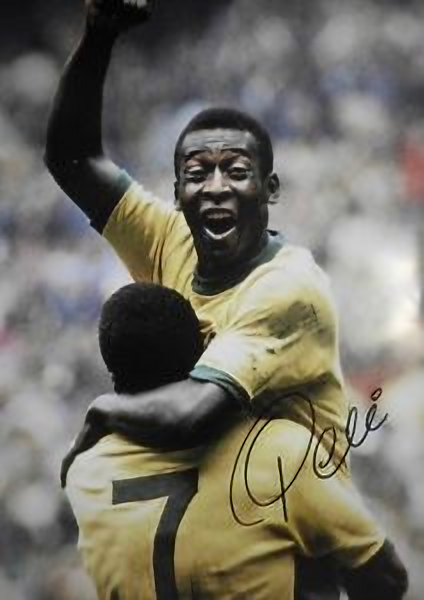}\\
    \caption{ZSSR}%
\vspace{2mm}
  \end{subfigure}%
\\
  \begin{subfigure}[b]{0.3\linewidth}
    \centering\includegraphics[width=150pt]{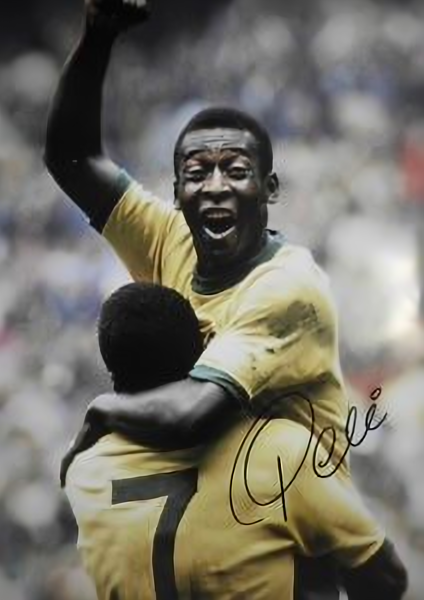}\\
    \caption{IDBP-CNN}%
\vspace{2mm}
  \end{subfigure}
  \begin{subfigure}[b]{0.3\linewidth}
    \centering\includegraphics[width=150pt]{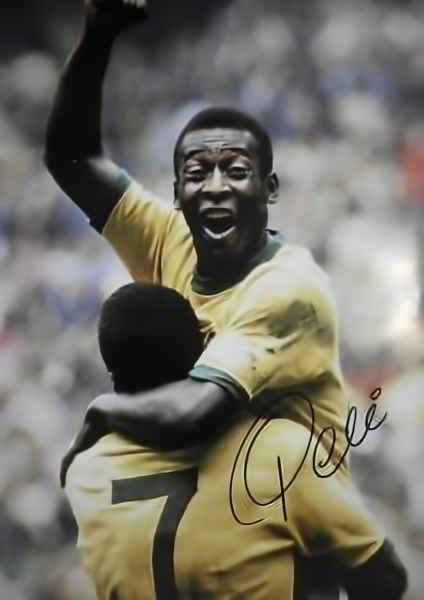}\\
    \caption{IDBP-CNN-IA}%
\vspace{2mm}
  \end{subfigure}
  \caption{SR(x2) of a real image. It is recommended to zoom at the images.}
\label{real_example4}
\end{figure*}

\begin{figure*}
\captionsetup[subfigure]{labelformat=empty}
  \centering
  \begin{subfigure}[b]{0.48\linewidth}
    \centering\includegraphics[width=75pt]{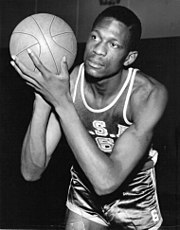}\\
    \caption{LR image}
\vspace{2mm}
  \end{subfigure}%
\\
  \begin{subfigure}[b]{0.3\linewidth}
    \centering\includegraphics[width=150pt]{img_024_SRF_2_EDSR}\\
    \caption{EDSR+}%
\vspace{2mm}
  \end{subfigure}%
  \begin{subfigure}[b]{0.3\linewidth}
    \centering\includegraphics[width=150pt]{img_024_SRF_2_RCAN}\\
    \caption{RCAN}%
\vspace{2mm}
  \end{subfigure}%
  \begin{subfigure}[b]{0.3\linewidth}
    \centering\includegraphics[width=150pt]{img_024_SRF_2_ZSSR}\\
    \caption{ZSSR}%
\vspace{2mm}
  \end{subfigure}%
\\
  \begin{subfigure}[b]{0.3\linewidth}
    \centering\includegraphics[width=150pt]{img_024_SRF_2_IDBPCNN}\\
    \caption{IDBP-CNN}%
\vspace{2mm}
  \end{subfigure}
  \begin{subfigure}[b]{0.3\linewidth}
    \centering\includegraphics[width=150pt]{img_024_SRF_2_ours}\\
    \caption{IDBP-CNN-IA}%
\vspace{2mm}
  \end{subfigure}
  \caption{SR(x2) of a real image. It is recommended to zoom at the images.}
\label{real_example5}
\end{figure*}

\begin{figure*}
\captionsetup[subfigure]{labelformat=empty}
  \centering
  \begin{subfigure}[b]{0.48\linewidth}
    \centering\includegraphics[width=75pt]{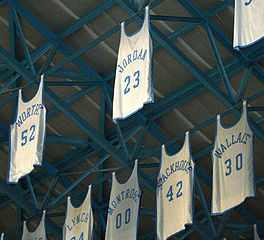}\\
    \caption{LR image}
\vspace{2mm}
  \end{subfigure}%
\\
  \begin{subfigure}[b]{0.3\linewidth}
    \centering\includegraphics[width=150pt]{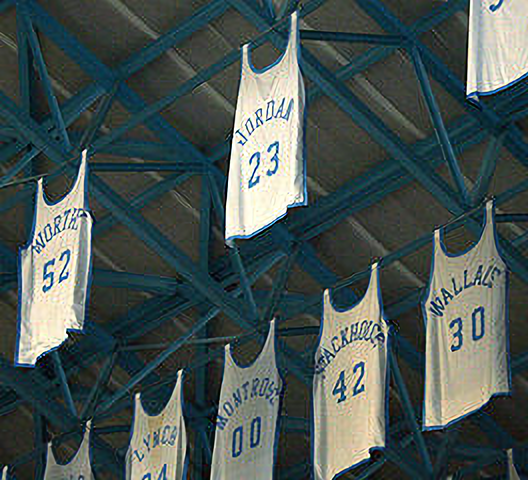}\\
    \caption{EDSR+}%
\vspace{2mm}
  \end{subfigure}%
  \begin{subfigure}[b]{0.3\linewidth}
    \centering\includegraphics[width=150pt]{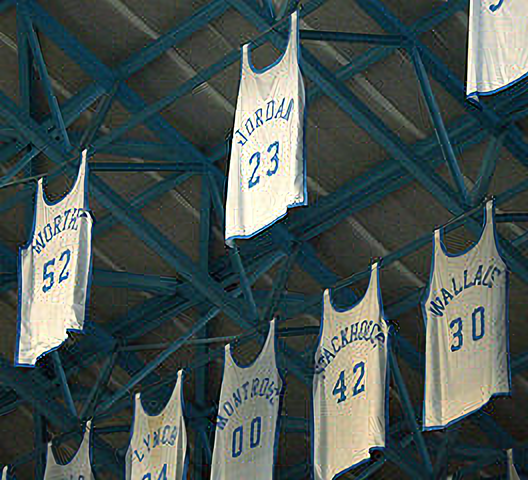}\\
    \caption{RCAN}%
\vspace{2mm}
  \end{subfigure}%
  \begin{subfigure}[b]{0.3\linewidth}
    \centering\includegraphics[width=150pt]{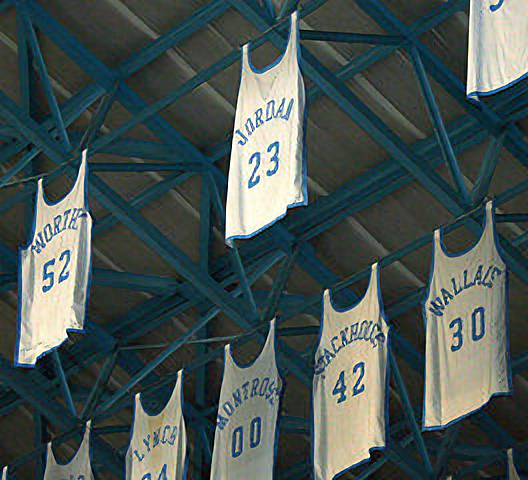}\\
    \caption{ZSSR}%
\vspace{2mm}
  \end{subfigure}%
\\
  \begin{subfigure}[b]{0.3\linewidth}
    \centering\includegraphics[width=150pt]{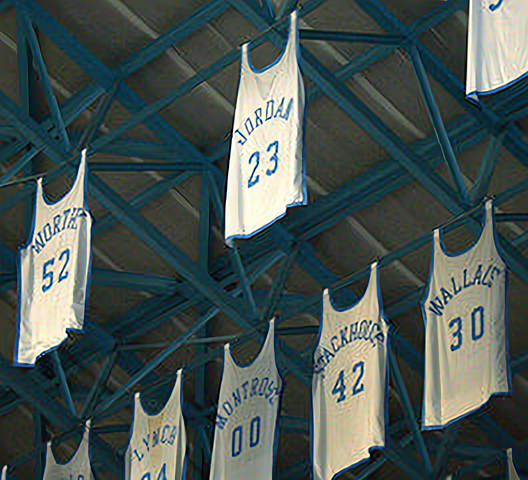}\\
    \caption{IDBP-CNN}%
\vspace{2mm}
  \end{subfigure}
  \begin{subfigure}[b]{0.3\linewidth}
    \centering\includegraphics[width=150pt]{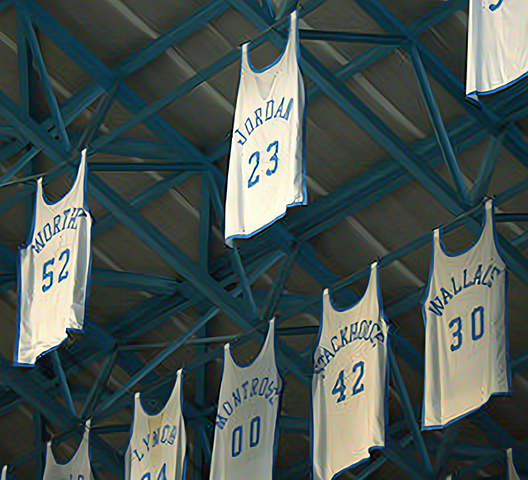}\\
    \caption{IDBP-CNN-IA}%
\vspace{2mm}
  \end{subfigure}
  \caption{SR(x2) of a real image. It is recommended to zoom at the images.}
\label{real_example6}
\end{figure*}

\begin{figure*}
\captionsetup[subfigure]{labelformat=empty}
  \centering
  \begin{subfigure}[b]{0.48\linewidth}
    \centering\includegraphics[width=75pt]{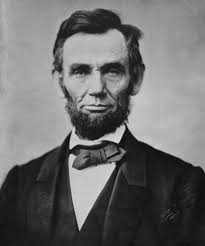}\\
    \caption{LR image}
\vspace{2mm}
  \end{subfigure}%
\\
  \begin{subfigure}[b]{0.3\linewidth}
    \centering\includegraphics[width=150pt]{img_002_SRF_2_EDSR}\\
    \caption{EDSR+}%
\vspace{2mm}
  \end{subfigure}%
  \begin{subfigure}[b]{0.3\linewidth}
    \centering\includegraphics[width=150pt]{img_002_SRF_2_RCAN}\\
    \caption{RCAN}%
\vspace{2mm}
  \end{subfigure}%
  \begin{subfigure}[b]{0.3\linewidth}
    \centering\includegraphics[width=150pt]{img_002_SRF_2_ZSSR}\\
    \caption{ZSSR}%
\vspace{2mm}
  \end{subfigure}%
\\
  \begin{subfigure}[b]{0.3\linewidth}
    \centering\includegraphics[width=150pt]{img_002_SRF_2_IDBPCNN}\\
    \caption{IDBP-CNN}%
\vspace{2mm}
  \end{subfigure}
  \begin{subfigure}[b]{0.3\linewidth}
    \centering\includegraphics[width=150pt]{img_002_SRF_2_ours}\\
    \caption{IDBP-CNN-IA}%
\vspace{2mm}
  \end{subfigure}
  \caption{SR(x2) of a real image. It is recommended to zoom at the images.}
\label{real_example7}
\end{figure*}

Figs. \ref{real_example} and \ref{real_example1}-\ref{real_example7} shows the reconstruction results of real old images. In all the examples our IDBP-CNN-IA technique clearly outperforms the other methods. 

\section{Conclusion}

Leading deep learning SR techniques are sensitive to the acquisition process assumptions used in the training phase. This work addressed this issue by combining two recent independent approaches, where the first solves inverse problems using existing denoisers and the second relies on internal information in the given LR image. 
Our contribution is using a fast internal learning step to fine-tune the CNN denoisers for the IDBP method, which is adapted here to the SR problem for the first time.
The proposed technique outperforms the above two approaches and gives better results than other state-of-the-art methods in practical cases, such as inexact downscaling kernels and low-quality real images.
Our proposed strategy is general and can be applied to other P\&P schemes as well.



\bibliographystyle{ieeetr}
\bibliography{paper_ver_final}

\end{document}